# SpectralDiT: Timestep-Conditioned Spectral Residual Correction for Flow-Matching DiTs

Jiayu Tian
Peking University

Code: github.com/lndezxn/SpectralDiT

## Abstract

We propose **SpectralDiT**, a lightweight modification to flow-matching Diffusion Transformers that adds timestep-conditioned spectral correction to the MLP residual branch. The module decomposes each residual update into low- and high-frequency components on the patch-token grid, then learns a zero-initialized additive gate so the model initially matches the baseline DiT. On CIFAR-10 pixel-space generation, SpectralDiT improves FID from 20.78 to 19.71 at patch size 1 and reduces the radial Fourier spectrum gap. Furthermore, we scale our method to latent diffusion on ImageNet-100. With $\sim$0.6% additional theoretical FLOPs and 1.36% additional parameters, SpectralDiT improves latent flow-matching, achieving an 8.7% relative FID reduction under classifier-free guidance (CFG 2.0). All reported results are averaged over five seeds. Ablations and gate visualizations on CIFAR-10 reveal stable block-specific spectral correction patterns.

## 1 Introduction

Diffusion Transformers have become a strong backbone for image generation [1], and flow-matching objectives further simplify training by learning continuous velocity fields from noise to data [2]. Although AdaLN-Zero modulates DiT residual branches with timestep and class information, the residual update itself is still applied uniformly across its spatial frequency components. This overlooks a simple but important observation: image generation is not spectrally uniform across time. Early stages tend to establish coarse structure, while later stages increasingly refine local variations and high-frequency details, echoing frequency-domain analyses of neural networks and diffusion models [3]; [4]; [5]; [6].

In this work, we ask whether this temporal spectral structure can be exploited with a minimal architectural change. We propose **SpectralDiT**, a timestep-conditioned spectral residual correction module for flow-matching DiTs. Instead of replacing the standard AdaLN-Zero pathway used in DiT [1], SpectralDiT preserves the original MLP residual branch and adds a zero-initialized correction term. Specifically, the MLP residual is decomposed on the patch-token grid into low- and high-frequency components using a fixed spatial low-pass filter. A lightweight gate, conditioned on the timestep embedding, then predicts how much each spectral component should contribute as an additive correction. Because the gate is zero-initialized, SpectralDiT initially matches the baseline DiT and learns to use spectral correction only when beneficial.

This design is intentionally small. It does not introduce new attention patterns, additional experts, auxiliary supervision, or a different sampling procedure. Instead, it tests a focused hypothesis: **can a DiT benefit from explicitly modulating the spectral composition of its residual updates over generation time?** This makes SpectralDiT easy to integrate, easy to ablate, and directly comparable to a standard flow-matching DiT [1]; [2].

We evaluate SpectralDiT on CIFAR-10 using the same flow-matching objective, sampler, training schedule, and model size as the baseline. Compared with the vanilla DiT at patch size 1, timestep-conditioned SpectralDiT improves FID from 20.78 to 19.71, increases recall from 0.559 to 0.576, and reduces the radial Fourier spectrum gap from 0.205 to 0.161 across five seeds. Analysis of the learned gates further reveals stable block-specific low-/high-frequency amplification and suppression patterns, supporting the view that residual updates in image generation benefit from explicit spectral specialization.

Our contributions are threefold:

1. We introduce a zero-initialized timestep-conditioned spectral residual correction module for flow-matching DiTs, adding $\sim$0.6% theoretical FLOPs and 1.36% parameters in our base configuration.
2. We provide a controlled comparison against standard DiT baselines under identical training and sampling settings. We analyze generated images through both distribution-level metrics and frequency/edge statistics, showing how spectral residual correction affects the generative process beyond FID alone.
3. We evaluate the scalability of our method on ImageNet-100 latent diffusion, where SpectralDiT improves FID, precision, and recall under Classifier-Free Guidance (CFG) across five seeds. We further verify the FID gains using a validation-only 5,000-image reference split, while spectrum-distance changes are reported as a diagnostic rather than a primary claim.

## 2 Related Work

**Diffusion and flow-based image generation.** Diffusion models have become a dominant framework for image synthesis by learning to reverse a gradual corruption process [7]. Subsequent work showed that architectural choices, guidance, and sampler design are critical for sample quality and efficiency [8]; [9]; [10]. Flow matching provides a closely related continuous-time formulation in which the model directly regresses a velocity field along probability paths between noise and data [2], and scalable interpolant transformers further connect transformer backbones with flow- and diffusion-style generative training [11]. Our work follows this line by using the same flow-matching objective and sampler for both the baseline and proposed model. Rather than changing the objective, guidance rule, or numerical solver, SpectralDiT studies whether a small architectural correction inside the residual pathway can better match the time-varying structure of the generative process.

**Transformer backbones for diffusion models.** Early high-performing diffusion models typically used U-Net backbones with convolutional inductive biases [7]; [8]. Vision Transformers process images as patch-token sequences [12], and Diffusion Transformers (DiTs) adapt this design to diffusion generation, showing strong scaling behavior as depth, width, and token count increase [1]. A key part of DiT is adaptive layer normalization with zero-initialized residual modulation, which makes conditional residual branches stable at the beginning of training [1]. SpectralDiT keeps this design intact: it does not introduce a new attention mechanism or a different tokenization scheme. Instead, it augments the MLP residual branch with a zero-initialized spectral correction, so the model initially matches the baseline DiT computation and learns only an additive low-/high-frequency adjustment.

**Frequency bias and spectral modeling.** A growing body of work shows that neural networks do not learn all frequencies uniformly. Spectral-bias analyses indicate that standard networks often fit lower-frequency structure faster than higher-frequency detail [3], while Fourier-feature mappings can improve the learning of high-frequency functions in coordinate-based networks [4]. Recent diffusion studies also show that frequency structure matters for both generation and sampling. DeCo decouples low-frequency semantics and high-frequency details in pixel diffusion [13], frequency-decoupled guidance separately controls low- and high-frequency CFG components during sampling [14], timestep-adaptive frequency enhancement improves diffusion-based super-resolution by modulating frequency components at different sampling stages [5], and decoupled diffusion transformers separate low-frequency semantic encoding from high-frequency decoding [15]. SpectralDiT differs in scope: rather than changing the guidance rule, adding a sampling-time frequency controller, or substantially redesigning the diffusion-transformer architecture, it inserts a zero-initialized architectural correction inside the DiT MLP residual branch. The correction is computed on the patch-token grid and is driven by the timestep embedding, directly testing whether the residual update itself benefits from learned low-/high-frequency modulation.

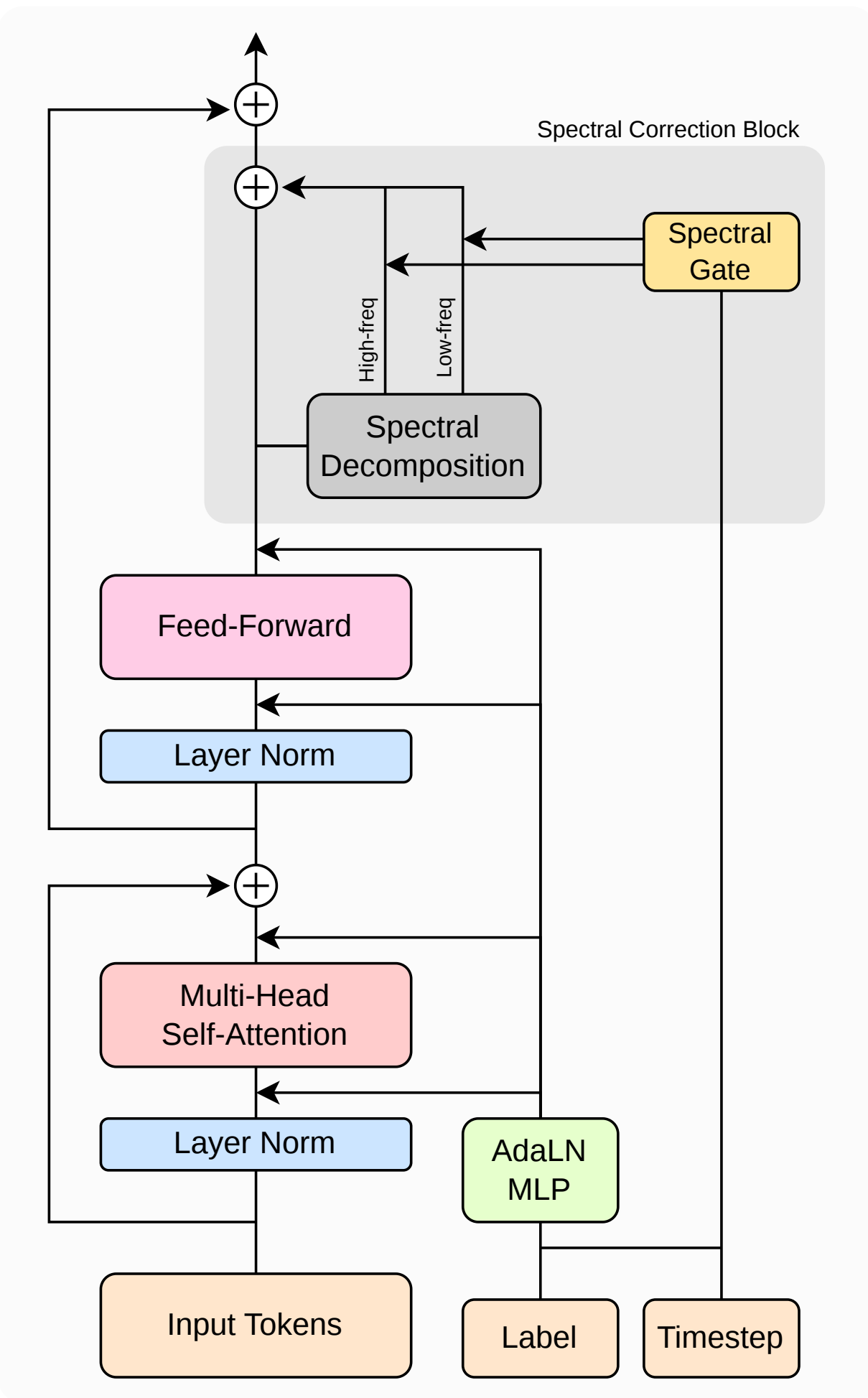

Figure 1: SpectralDiT architecture overview. The spectral correction block is added in the MLP residual branch of each DiT block.

## 3 Method

### 3.1 Overview

We propose **SpectralDiT**, a minimal modification to flow-matching Diffusion Transformers that adds a timestep-conditioned spectral correction to the MLP residual branch of each DiT block. The key idea is to preserve the original AdaLN-Zero DiT pathway while allowing the model to learn additive low-/high-frequency residual corrections over generation time.

Given an input image or latent state, the model first follows a standard flow-matching DiT forward pass. In each Transformer block, after the AdaLN-Zero-modulated MLP residual is computed, we decompose this residual on the patch-token grid into low- and high-frequency components. A lightweight gate then predicts two scalar coefficients controlling how much low- and high-frequency correction should be added back. The gate can be static, timestep-conditioned, or timestep-and-class-conditioned. Our main model uses the timestep-conditioned version. The pipeline is illustrated in Figure 1.

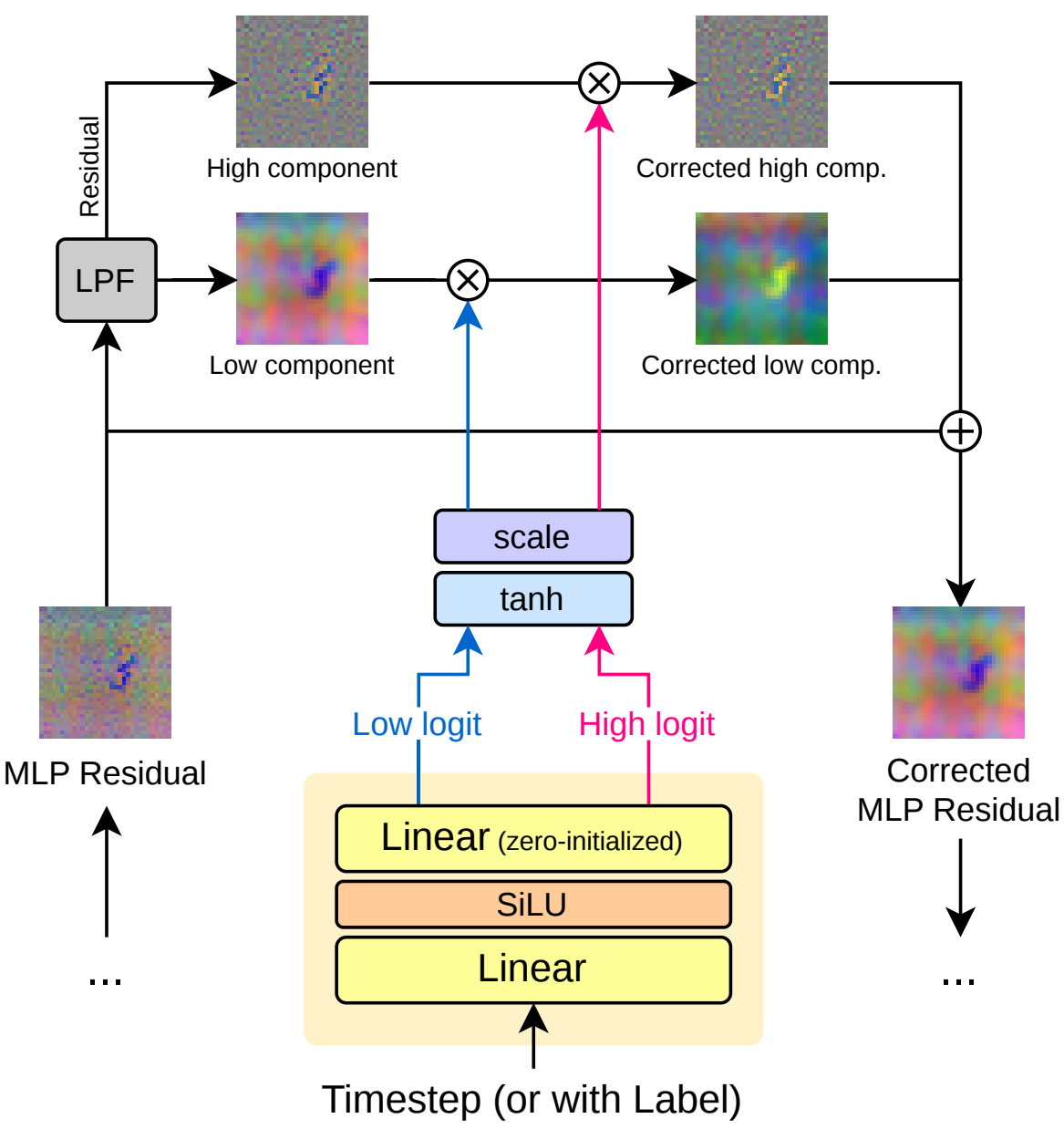

Figure 2: Detail of the spectral residual correction block. The MLP residual is decomposed into low- and high-frequency components using a fixed low-pass filter. A zero-initialized gate predicts additive correction weights for each component.

## 3.2 Baseline Flow-Matching DiT

We use a class-conditional DiT as the baseline backbone. Following patch-based vision transformers [12] and DiT [1], the input $x_t \in \mathbb{R}^{C \times H \times W}$ is divided into non-overlapping patches and embedded into a token sequence $z \in \mathbb{R}^{N \times d}$, where $N = H' \times W'$ is the number of patch tokens and $d$ is the hidden dimension. A fixed 2D sinusoidal positional embedding is added to the token sequence.

Each DiT block uses AdaLN-Zero conditioning, which predicts shift, scale, and gate parameters for the attention and MLP residual branches. Given the timestep embedding $e_t$ and class embedding $e_y$, the AdaLN-Zero parameters are computed as:

$$(\Delta_{\text{msa}}, \Gamma_{\text{msa}}, g_{\text{msa}}, \Delta_{\text{mlp}}, \Gamma_{\text{mlp}}, g_{\text{mlp}}) = \text{AdaLN}(e_t + e_y)$$

For an input token sequence $z$, the attention and MLP branch outputs are computed as:

$$r_{\text{msa}} = g_{\text{msa}} \odot \text{MSA}(\text{LN}(z) \odot (1 + \Gamma_{\text{msa}}) + \Delta_{\text{msa}})$$
$$z' = z + r_{\text{msa}}$$
$$r_{\text{mlp}} = g_{\text{mlp}} \odot \text{MLP}(\text{LN}(z') \odot (1 + \Gamma_{\text{mlp}}) + \Delta_{\text{mlp}})$$
$$z_{\text{out}} = z' + r_{\text{mlp}}$$

Our method modifies only the final MLP residual update.

## 3.3 Spectral Residual Correction

The central operation of SpectralDiT is applied to the MLP residual $r_{\text{mlp}}$. Since DiT operates on patch tokens, we interpret the token sequence as a 2D patch-token grid. Given $r_{\text{mlp}} \in \mathbb{R}^{N \times d}$, we reshape it into $R_{\text{mlp}} \in \mathbb{R}^{d \times H' \times W'}$.

We then compute a fixed low-pass filtered component $R_{\text{low}}$ and a high-frequency component $R_{\text{high}}$ using a depthwise spatial filter, following the low-pass/residual decomposition used in image pyramids [16]. In our implementation, we use the 1D kernel $k = \frac{1}{16}[1, 4, 6, 4, 1]$ and construct the 2D kernel by the outer product. The low-pass filter is applied separately to each channel with replicate padding at the spatial boundary, and the high-frequency component is obtained by subtracting the low-pass component from the original:

$$R_{\text{low}} = \text{LPF}(R_{\text{mlp}})$$
$$R_{\text{high}} = R_{\text{mlp}} - R_{\text{low}}$$

A zero-initialized frequency gate $G$ predicts two scalar correction coefficients:

$$(\delta_{\text{low}}, \delta_{\text{high}}) = s \cdot \tanh(G(\cdot))$$

The corrected MLP residual is then:

$$R'_{\text{mlp}} = R_{\text{mlp}} + \delta_{\text{low}} R_{\text{low}} + \delta_{\text{high}} R_{\text{high}}$$

After reshaping $R'_{\text{mlp}}$ back into token form $r'_{\text{mlp}}$, the block output becomes:

$$z_{\text{out}} = z' + r'_{\text{mlp}}$$

The process above is illustrated in Figure 2. For visualization, we project the hidden feature map to RGB by grouping channels according to their channel index modulo 3. Channels with indices $c \bmod 3 = 0, 1, 2$ are assigned to the R, G, and B channels, respectively, and each group is averaged before normalization for display. This projection is used only for visualization and does not correspond to physical RGB colors.

For the 8-block hidden-size-192 model, SpectralDiT adds 74,896 parameters, corresponding to 1.36% overhead.

## 3.4 Gate Variants

### Static Gate

The static variant uses block-specific learned scalars that do not depend on timestep or class label:

$$(\delta_{\text{low}}, \delta_{\text{high}}) = s \cdot \tanh(a_l)$$

where $a_l \in \mathbb{R}^2$ is a learned parameter for block $l$.

### Timestep-Conditioned Gate

Our main model uses a timestep-conditioned gate. For each block $l$, the gate is predicted from the timestep embedding:

$$(\delta_{\text{low}}, \delta_{\text{high}}) = s \cdot \tanh(G_l(e_t))$$

where $G_l$ is a small two-layer MLP:

$$G_{l(e_t)} = W_{2,l} \cdot \text{SiLU}(W_{1,l} e_t)$$

This version directly tests our main hypothesis: different generation times may benefit from different spectral corrections.

**Timestep-and-Class-Conditioned Gate**
An extended variant conditions the gate on both timestep and class label:

$$(\delta_{\text{low}}, \delta_{\text{high}}) = s \cdot \tanh\big(G_l(e_t + e_y)\big)$$

Unless otherwise noted, **SpectralDiT** refers to the timestep-conditioned version.

### 3.5 Zero-Initialized Additive Correction

A key design choice is that the spectral branch is initialized as an additive correction rather than a replacement. The final linear layer of the gate is initialized to zero, so SpectralDiT is initially identical to the baseline DiT. This avoids destabilizing the AdaLN-Zero training dynamics and ensures that any later deviation from the baseline is learned only when beneficial.

The spectral gate is therefore not intended to replace the original AdaLN-Zero gates. Instead, it provides an additional low-/high-frequency correction on top of the standard MLP residual.

### 3.6 Computational Efficiency and Parameter Overhead

SpectralDiT is designed to be highly parameter-efficient. The spectral decomposition relies on fixed, depthwise low-pass filtering with no learnable weights. The only newly introduced parameters are the lightweight MLP gates predicting the scalar coefficients. For our base DiT-S/2 configuration, this module introduces approximately 1.36% additional parameters and 0.6% additional theoretical FLOPs.

### 3.7 Strict Orthogonality to Macro-Level Spectral Decoupling

SpectralDiT is strictly orthogonal to macro-level spectral decoupling methods such as DeCo [13] and DDT [15]. Those methods alter how semantic low-frequency and detail high-frequency information are represented, routed, or decoded at the architecture or pipeline level. In contrast, SpectralDiT leaves the DiT tokenization, attention pathway, objective, conditioning pathway, sampler, and decoder unchanged. Its only operation is a local residual-block transformation:

$$r_{\text{mlp}} \to r_{\text{mlp}} + \delta_{\text{low}} R_{\text{low}} + \delta_{\text{high}} R_{\text{high}}$$

applied after the standard AdaLN-Zero-modulated MLP branch. Mathematically, this is a micro-level reweighting of the already-computed residual update; architecturally, it is a plug-in inside each Transformer block rather than a replacement for the model's global frequency decomposition strategy.

Therefore, SpectralDiT and macro-level decoupling approaches do not define competing alternatives. A model can first adopt DeCo-style or DDT-style global low-/high-frequency separation and still insert SpectralDiT inside its Transformer residual blocks, because the former decides how frequency information is organized across the generative architecture while the latter decides how each local residual update is spectrally corrected over time. The relevant controlled comparison for this paper is thus the DiT backbone with and without this residual correction under identical training and sampling protocols, rather than an external leaderboard-style comparison against systems that modify different parts of the generative stack.

## 4 Experiments

### 4.1 Experimental Setup

To evaluate the efficacy of the proposed spectral residual correction, we conduct controlled image-generation experiments across two distinct paradigms: pixel-space generation on CIFAR-10 [17] and latent-space generation on ImageNet-100. Our experimental philosophy prioritizes isolating the structural impact of SpectralDiT under strictly identical training and sampling configurations against standard flow-matching DiT baselines [1]; [2].

**Benchmark Settings**
**CIFAR-10 (Pixel-Space Core).** For the fundamental validation and mechanistic analysis, we train a full-resolution DiT (Patch Size 1, 1024 tokens) as well as coarser variants (Patch Sizes 2 and 4) at $32 \times 32$ resolution. Images are normalized to $[-1, 1]$ without data augmentation.

**ImageNet-100 (Latent-Space Scalability).** To test whether the same architectural modification transfers beyond pixel-space CIFAR-10, we evaluate on ImageNet-100 at $256 \times 256$ resolution. Following latent diffusion models [18], images are mapped to a $32 \times 32$ latent space using the public pre-trained stabilityai/sd-vae-ft-ema VAE. We evaluate both unguided sampling (CFG 1.0) and guided sampling (CFG 2.0).

**Training and Sampling Protocols**
All models are trained utilizing a linear flow-matching objective for 400,000 steps with an AdamW optimizer [19] (batch size of 256, $\beta_1 = 0.9$, $\beta_2 = 0.99$, no weight decay). The learning rate follows a cosine schedule starting at 0.0001 after 5,000 warmup steps, down to a minimum of 0.00001. We enforce gradient clipping at 1.0 and maintain an exponential moving average (EMA) of model weights with a decay rate of 0.9999. Classifier-free dropout probability is set to 0.1 during training. All inference sampling utilizes standard Euler integration with 50 deterministic steps. Experiments are run on a single NVIDIA GeForce RTX 5090 GPU. Unless otherwise noted, all quantitative results are averaged over five random seeds (0, 1, 2, 3, 4).

**Computational and Parameter Efficiency**
SpectralDiT is designed to be extremely lightweight from a parameter perspective. The spatial decomposition relies entirely on fixed, depthwise 1D low-pass kernels applied as separable 2D convolutions, introducing zero learnable parameters. The only trainable overhead stems from the miniature two-layer MLP gating modules. As detailed in Table 1, for the base 8-block DiT-S/2 backbone, SpectralDiT adds a mere 74,896 parameters and increases theoretical FLOPs by approximately 0.6%.

| Patch size | 1 / 2 / 4 |
|---|---|
| Hidden dimension $d$ | 192 |
| Gate hidden dimension | 48 |
| Gate scale factor ($s$) | 1.0 |
| Attention heads | 4 |
| MLP ratio | 4 |
| DiT Block count | 8 |
| Learnable Parameter Overhead | +74,896 ( 1.36%) |
| Theoretical FLOPs Overhead | 0.6% |

Table 1: Architectural and computational configuration details.

## 4.2 Evaluation Metrics

We deploy a multi-faceted evaluation protocol spanning distribution-level fidelity, sample diversity, and precise frequency-domain alignment:

- **Fréchet Inception Distance (FID)** [20]: Measures overall generative distribution distance. We report FID-10K for CIFAR-10 and FID-50K for the primary ImageNet-100 comparison. For ImageNet-100, the 50,000-image reference statistics are computed from a randomly sampled mixed real-image pool to provide a statistically stable, low-variance internal comparison between architectures under the same evaluation protocol. To rule out potential train-set leakage concerns, we additionally report FID-5K computed strictly against the 5,000-image ImageNet-100 validation split.
- **Inception Score (IS)**: Evaluates the conditional specificity and diversity of generated samples.
- **Precision and Recall** [21]: Deconstructed in Inception-v3 feature space (neighborhood size 3) to explicitly decouple sample fidelity (Precision) from distribution coverage (Recall).
- **Radial Fourier Spectrum Distance ($D_{\text{spectrum}}$)**: To quantify mathematical frequency alignment [22], we compute the 2D Fourier magnitude spectrum for each image, average the energy across radial bins $r$, and calculate the mean log-absolute discrepancy against real images:

$$D_{\text{spectrum}} = \frac{1}{R}\sum_{r=1}^{R} |\log S_{\text{real}}(r) - \log S_{\text{gen}}(r)|$$

- **Spatial Edge Statistics**: We report absolute gaps in dataset-level mean and standard deviation for Sobel gradient magnitudes, alongside mean absolute Laplacian ($\nabla^2$) response gaps, characterizing sharpness alignment.

# 5 Results

## 5.1 Core Generation Quality on CIFAR-10

| Patch | Gating Condition | FID-10K ↓ | Precision ↑ | Recall ↑ |
|---|---|---|---|---|
| 1 | None (Vanilla DiT) | 20.78 | **0.610** | 0.559 |
| 1 | Static Gate | 20.60 | 0.609 | 0.563 |
| 1 | Timestep-Cond. (Ours) | **19.71** | 0.603 | **0.576** |
| 1 | Time & Class Cond. | 19.90 | 0.602 | 0.574 |
| 2 | None (Vanilla DiT) | 23.99 | **0.595** | 0.550 |
| 2 | Timestep-Cond. (Ours) | **23.17** | 0.593 | **0.556** |
| 4 | None (Vanilla DiT) | **34.58** | **0.591** | **0.467** |
| 4 | Timestep-Cond. (Ours) | 34.90 | 0.589 | 0.457 |

Table 2: Pixel-space generation performance on CIFAR-10. Values are means over seeds 0–4; standard deviations are summarized in the surrounding text to keep the table compact. Timestep-conditioned SpectralDiT yields the best FID and recall at fine token resolutions, while performance degrades at excessive patch sizes.

Table 2 presents the controlled comparison on CIFAR-10 across varied token-grid resolutions and gating formulations. At Patch Size 1, our timestep-conditioned SpectralDiT establishes a clear advantage, dropping FID from 20.78 to 19.71 (a 5.2% relative enhancement). This improvement is accompanied by a boost in Recall from 0.559 to 0.576, while Precision remains competitive. The combination suggests that the structural spectral modulation improves distribution coverage rather than merely acting as an aggressive sharpening heuristic.

Because the main tables are space-constrained, we report uncertainty in text rather than inside each cell. For the central patch-size-1 comparison, the FID change is 20.78 +/- 0.24 for vanilla DiT versus 19.71 +/- 0.17 for SpectralDiT, and Recall changes from 0.559 +/- 0.005 to 0.576 +/- 0.004 across five seeds. Thus, the main CIFAR-10 FID and Recall improvements are larger than the observed run-to-run variation.

Analyzing gate variants supports our core conceptual hypothesis: the Static Gate yields only a small improvement in FID (20.78 to 20.60), while the timestep-conditioned gate performs substantially better (19.71). Introducing Class Conditioning alongside Timestep info (Time & Class) similarly underperforms the pure Timestep variant (19.90 vs. 19.71), indicating that the dynamic spectral demand of velocity fields is primarily a function of generation time, whereas auxiliary conditioning signals can over-complicate the learned gating profiles.

Furthermore, the performance gains depend on token-grid spatial resolution. SpectralDiT remains effective at Patch Size 2 (FID 23.99 to 23.17), but its advantage vanishes at Patch Size 4 (34.58 to 34.90). This behavior aligns with signal processing intuition: an excessively coarse token grid ($8 \times 8$ at $p = 4$) undergoes severe spatial aliasing, rendering token-level frequency decomposition structurally less informative for residual modifications.

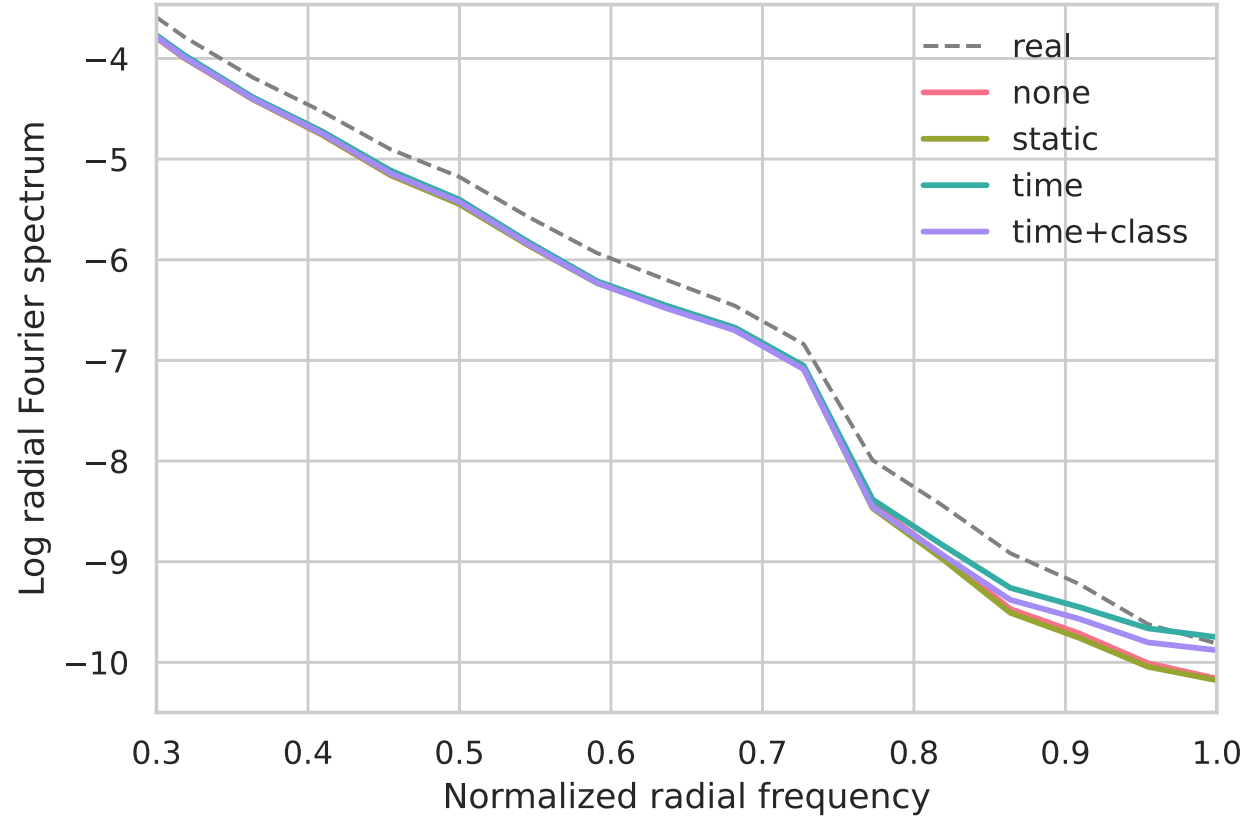

Figure 3: Radial Fourier spectrum profiles at patch size 1. SpectralDiT closely tracks the real data distribution, particularly mitigating the high-frequency deficit endemic to baseline architectures.

## 5.2 Frequency Alignment and Edge Profiles

| Patch | Condition | $D_{\text{spectrum}} \downarrow$ | Sobel Mean/Std $\Delta \downarrow$ | $\nabla^2\Delta \downarrow$ |
|---|---|---|---|---|
| 1 | None | 0.205 | 0.031 / 0.034 | 0.0123 |
| 1 | Static | 0.196 | 0.030 / 0.033 | 0.0119 |
| 1 | Timestep (Ours) | **0.161** | **0.029 / 0.033** | **0.0107** |
| 1 | Time & Class | 0.178 | 0.032 / 0.033 | 0.0124 |

Table 3: Spectral and spatial edge-statistic discrepancies at patch size 1 on CIFAR-10. Values are means over seeds 0–4; standard deviations for the key spectrum-distance comparison are summarized in the text.

To clarify the physical mechanism driving the FID drops, we evaluate frequency-space alignment in Table 3 and Figure 3. Timestep-conditioned SpectralDiT achieves a substantial reduction in Radial Fourier Spectrum Distance, falling from 0.205 +/- 0.004 to 0.161 +/- 0.002. As illustrated by the full spectrum curves in Figure 3, while all deep generative models inherently exhibit an energy deficit in extreme high-frequency bands, SpectralDiT mitigates this phenomenon, aligning more closely with real data statistics. Spatial edge metrics mirror this trend: the Sobel and Laplacian gaps are also lower for the timestep-conditioned variant. This indicates that the learned gate does not simply execute naive edge-sharpening; instead, it improves the alignment of the natural image frequency spectrum.

## 5.3 Scalability to Latent Diffusion: ImageNet-100

To evaluate whether the same idea remains useful in a latent diffusion setting, we scale SpectralDiT to latent-space flow-matching on ImageNet-100. We also benchmark the architecture under Classifier-Free Guidance (CFG), a standard inference technique that often improves fidelity while reducing diversity. The primary 50,000-sample results are reported in Table 4, and validation-only FID-5K results are reported in Table 5.

| CFG | Model | FID-50K ↓ | IS ↑ | Precision ↑ | Recall ↑ | $D_{\text{spectrum}} \downarrow$ |
|---|---|---|---|---|---|---|
| 1.0 | DiT-S/2 | 62.70 | 22.33 | 0.296 | **0.556** | 0.700 |
| 1.0 | **SpectralDiT** | **59.03** | **23.53** | **0.310** | 0.556 | **0.693** |
| 2.0 | DiT-S/2 | 14.88 | 53.80 | 0.593 | 0.341 | 0.697 |
| 2.0 | **SpectralDiT** | **13.59** | **55.80** | **0.605** | **0.347** | **0.693** |

Table 4: Latent image-generation metrics on ImageNet-100 ($256 \times 256$). Values are means over seeds 0–4; standard deviations are summarized in the text. The 50,000-image reference set is used for a statistically stable internal comparison; strictly validation-only FID is reported separately in Table 5.

| CFG | Model | Val FID-5K ↓ | Std. |
|---|---|---|---|
| 1.0 | DiT-S/2 | 68.34 | 0.70 |
| 1.0 | **SpectralDiT** | **64.30** | 0.60 |
| 2.0 | DiT-S/2 | 25.41 | 0.24 |
| 2.0 | **SpectralDiT** | **24.49** | 0.15 |

Table 5: Validation-only FID on ImageNet-100. The reference set is restricted to the 5,000-image validation split, eliminating possible train-set leakage in the FID reference statistics. Values are means and standard deviations over seeds 0–4.

The latent-space experiments reveal two major phenomena. We use a 50,000-image reference set in Table 4 to obtain a statistically stable and low-variance FID estimate, while Table 5 validates the same architectural comparison strictly on the 5,000-image validation split to eliminate potential train-set leakage concerns. First, in the unconditional regime (CFG 1.0), SpectralDiT transfers to the latent setting under the same training and sampling protocol, reducing FID-50K from 62.70 +/- 0.14 to 59.03 +/- 0.16 and improving Inception Score from 22.33 to 23.53. Under the validation-only protocol, the corresponding FID-5K also improves from 68.34 +/- 0.70 to 64.30 +/- 0.60. Precision also improves from 0.296 to 0.310, while Recall remains essentially unchanged at 0.556 for both models.

Second, SpectralDiT also improves the guided setting (CFG 2.0). When CFG is applied, the baseline DiT's FID-50K drops sharply to 14.88 while Recall decreases from 0.556 to 0.341, consistent with the expected fidelity-diversity pressure of guidance. In this guided regime, SpectralDiT pushes the FID-50K further down from 14.88 +/- 0.10 to 13.59 +/- 0.04 (an 8.7% relative gain), improves Inception Score from 53.80 to 55.80, and reaches a higher Precision of 0.605. The validation-only FID-5K comparison shows the same direction, improving from 25.41 +/- 0.24 to 24.49 +/- 0.15. Recall also improves from 0.341 +/- 0.002 to 0.347 +/- 0.003.

This behavior suggests that timestep-conditioned spectral correction may act as an internal architectural regularizer. The ImageNet-100 spectrum-distance results are more cautious than the FID results: under CFG 1.0, the spectrum distance changes from 0.700 +/- 0.001 to 0.693 +/- 0.001, while under CFG 2.0 it changes only from 0.697 +/- 0.002 to 0.693 +/- 0.005. We therefore interpret the ImageNet-100 spectrum metric as a supporting diagnostic rather than strong evidence of high-frequency alignment. Together with the FID, precision, and recall gains, the results support the narrower claim that residual frequency gating improves sample quality under this controlled latent flow-matching setup.

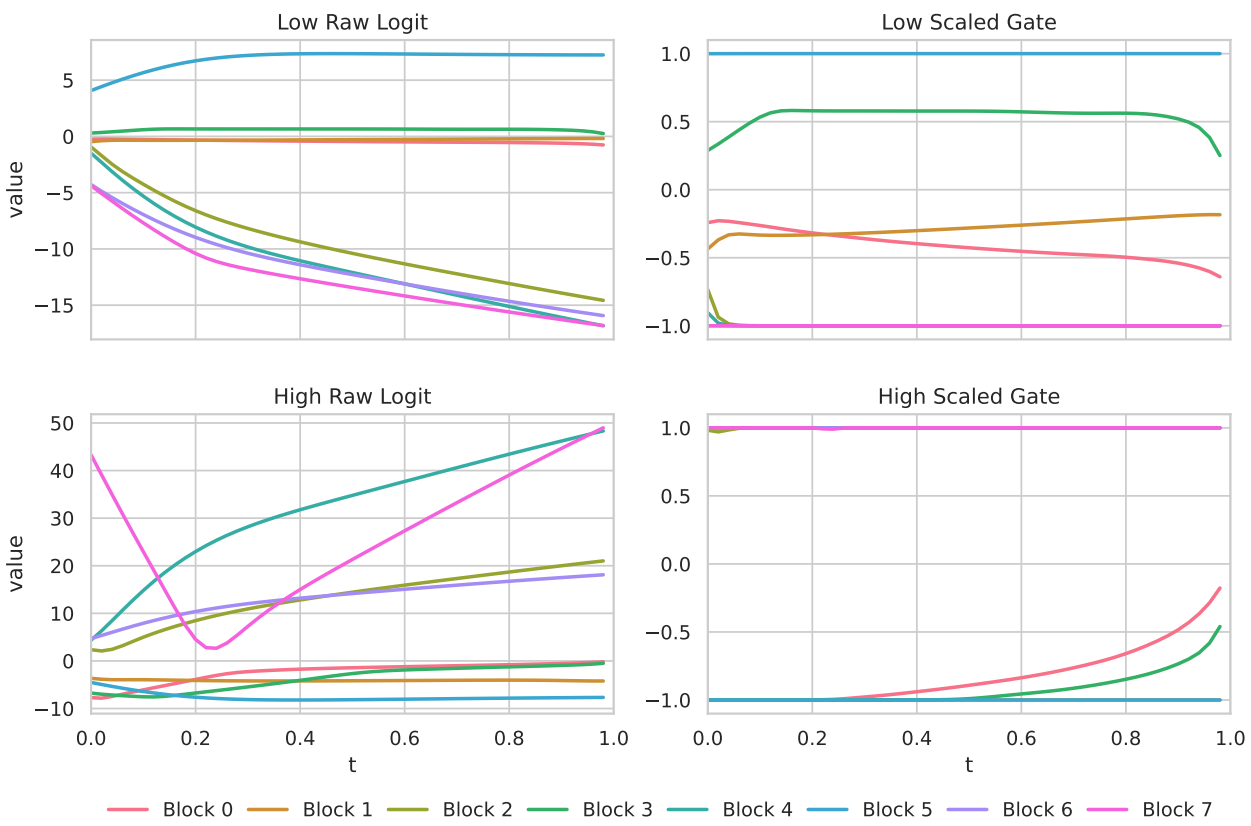

Figure 4: Learned spectral gate trajectories for the timestep-conditioned SpectralDiT model ($p = 1,\ s = 1.0$). Curves track individual Transformer blocks across generation time $t$. Saturated bounds indicate strong block-specific frequency correction.

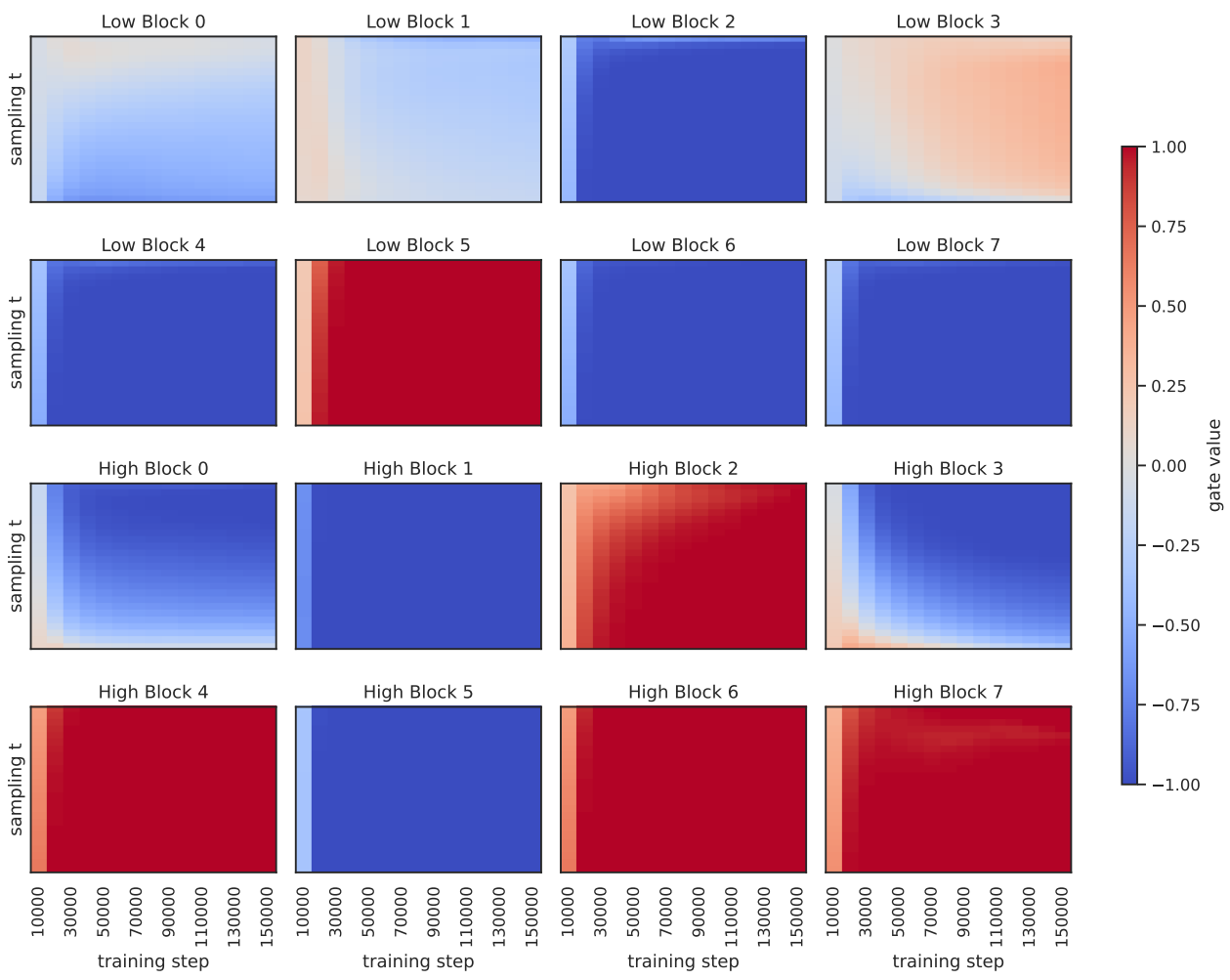

Figure 5: Chronological training evolution of spectral gate weights ($p = 1,\ s = 1.0$). Polarization away from zero initialization indicates active, non-trivial optimization.

**Online Convergence and Latency Trade-offs**

The performance gains are also visible in the training logs: at 400,000 steps, the online evaluation registers a baseline FID of 64.78 (IS 21.67) versus SpectralDiT's FID of 61.39 (IS 22.90), suggesting a faster convergence trend in this run.

In terms of real-world deployment, our unoptimized PyTorch implementation incurs a practical inference wall-clock latency overhead despite the small theoretical FLOPs increase. In the ImageNet-100 50-step evaluation, unconditional (CFG 1.0) per-sample generation time rises from 0.00456 seconds (Baseline) to 0.00493 seconds (SpectralDiT); under CFG 2.0, the times are 0.00886 and 0.00952 seconds, respectively. We attribute much of this latency to memory access and tensor layout overhead from frequent reshaping (`flatten`, `view`, `transpose`) required to enforce grid contiguity for spatial filtering. Given that our theoretical FLOPs increment is approximately $\sim$0.6% and learnable parameter growth is 1.36%, the result suggests a useful accuracy-efficiency trade-off, although optimized kernels would be needed to realize the theoretical overhead in wall-clock inference.

## 5.4 Gate Behavior Analysis

To interpret the internal mechanics of SpectralDiT, we extract and visualize the learned frequency gate trajectories ($\delta_{\text{low}}$, $\delta_{\text{high}}$) from the CIFAR-10 model. Bound within $[-1, 1]$ via the scaled $s \cdot \tanh(\cdot)$ formulation, our additive correction mathematically mutates the standard MLP residual step into a weighted combination:

$$R'_{\text{mlp}} = (1 + \delta_{\text{low}})R_{\text{low}} + \big(1 + \delta_{\text{high}}\big)R_{\text{high}}$$

Consequently, a gate approaching $+1$ doubles the representation intensity of that specific frequency band, whereas $-1$ acts as a complete suppression filter.

As demonstrated in Figure 4, the model undergoes significant polarization. Rather than settling into minor, perturbative adjustments, the raw logits drive multiple blocks toward the extreme boundaries ($+1$ or $-1$). This explicit utilization indicates that SpectralDiT is not merely an unused auxiliary branch; it learns strong block-wise frequency corrections.

These behaviors are also stratified by block layer depth. Instead of imposing a homogeneous spectral shift across the whole network, different Transformer blocks learn different correction profiles: some intermediate blocks amplify low-frequency components while suppressing high-frequency residuals, whereas later blocks more strongly scale high-frequency residual components.

The chronological emergence of this behavior is captured in Figure 5. Initiated at zero to preserve AdaLN-Zero training dynamics, the gates differentiate during training and settle into stable temporal profiles. Interestingly, while the gates possess the architectural freedom to modulate smoothly across timesteps, several blocks converge to near-constant saturation lines across the generation path, while a subset retains finer timestep variation. This suggests that SpectralDiT combines layer-wise spectral specialization with localized timestep-adaptive frequency modulation.

# 6 Limitations

While SpectralDiT introduces small theoretical FLOPs and parameter overhead, the explicit token-to-grid reshaping currently incurs a noticeable inference latency penalty due to memory access costs. Future work could address this by developing fused custom CUDA kernels for patch-aware spectral filtering. In addition, although we include validation-only FID-5K results to address train-set leakage concerns in the ImageNet-100 reference statistics, broader datasets, larger validation protocols, and evaluation on billion-parameter text-to-image foundation models remain important next steps.

# 7 Conclusion

This work presented **SpectralDiT**, a zero-initialized timestep-conditioned spectral correction for the MLP residual branch of flow-matching DiTs. By separating residual updates into low- and high-frequency components, the model can learn block-specific spectral adjustments while preserving the baseline computation at initialization.

On CIFAR-10 pixel-space generation, SpectralDiT improves FID and Fourier spectrum alignment across five seeds, with positive recall changes in the strongest settings. On ImageNet-100 latent-space generation, it improves FID, precision, and recall under the controlled 50,000-image evaluation protocol, and the FID trend persists under a strict validation-only FID-5K protocol. Spectrum-distance changes should be interpreted cautiously. These results support the view that explicit residual frequency modulation is a promising architectural inductive bias for modern diffusion transformers, while motivating broader large-scale validation.

# References

[1] W. Peebles and S. Xie, "Scalable Diffusion Models with Transformers," in *Proceedings of the IEEE/CVF International Conference on Computer Vision*, 2023.

[2] Y. Lipman, R. T. Q. Chen, H. Ben-Hamu, M. Nickel, and M. Le, "Flow Matching for Generative Modeling," in *International Conference on Learning Representations*, 2023.

[3] N. Rahaman *et al.*, "On the Spectral Bias of Neural Networks," in *Proceedings of the 36th International Conference on Machine Learning*, in Proceedings of Machine Learning Research, vol. 97. 2019, pp. 5301–5310.

[4] M. Tancik *et al.*, "Fourier Features Let Networks Learn High Frequency Functions in Low Dimensional Domains," in *Advances in Neural Information Processing Systems*, 2020.

[5] Y. Li *et al.*, "A Timestep-Adaptive Frequency-Enhancement Framework for Diffusion-based Image Super-Resolution," in *Proceedings of the Thirty-Fourth International Joint Conference on Artificial Intelligence*, 2025, pp. 1503–1511.

[6] F. Falck *et al.*, "A Fourier Space Perspective on Diffusion Models," *arXiv preprint arXiv:2505.11278*, 2025.

[7] J. Ho, A. Jain, and P. Abbeel, "Denoising Diffusion Probabilistic Models," in *Advances in Neural Information Processing Systems*, 2020.

[8] P. Dhariwal and A. Nichol, "Diffusion Models Beat GANs on Image Synthesis," in *Advances in Neural Information Processing Systems*, 2021.

[9] J. Ho and T. Salimans, "Classifier-Free Diffusion Guidance," *arXiv preprint arXiv:2207.12598*, 2022.

[10] T. Karras, M. Aittala, T. Aila, and S. Laine, "Elucidating the Design Space of Diffusion-Based Generative Models," in *Advances in Neural Information Processing Systems*, 2022.

[11] N. Ma, M. Goldstein, M. S. Albergo, N. M. Boffi, E. Vanden-Eijnden, and S. Xie, "SiT: Exploring Flow and Diffusion-Based Generative Models with Scalable Interpolant Transformers," in *European Conference on Computer Vision*, 2024.

[12] A. Dosovitskiy *et al.*, "An Image is Worth 16x16 Words: Transformers for Image Recognition at Scale," in *International Conference on Learning Representations*, 2021.

[13] Z. Ma, L. Wei, S. Wang, S. Zhang, and Q. Tian, "DeCo: Frequency-Decoupled Pixel Diffusion for End-to-End Image Generation," *arXiv preprint arXiv:2511.19365*, 2025.

[14] S. Sadat, T. Vontobel, F. Salehi, and R. M. Weber, "Guidance in the Frequency Domain Enables High-Fidelity Sampling at Low CFG Scales," *arXiv preprint arXiv:2506.19713*, 2025.

[15] S. Wang, Z. Tian, W. Huang, and L. Wang, "DDT: Decoupled Diffusion Transformer," *arXiv preprint arXiv:2504.05741*, 2025.

[16] P. J. Burt and E. H. Adelson, "The Laplacian Pyramid as a Compact Image Code," *IEEE Transactions on Communications*, vol. 31, no. 4, pp. 532–540, 1983.

[17] A. Krizhevsky, "Learning Multiple Layers of Features from Tiny Images," technical report, 2009.

[18] R. Rombach, A. Blattmann, D. Lorenz, P. Esser, and B. Ommer, "High-Resolution Image Synthesis with Latent Diffusion Models," in *Proceedings of the IEEE/CVF Conference on Computer Vision and Pattern Recognition*, 2022, pp. 10684–10695.

[19] I. Loshchilov and F. Hutter, "Decoupled Weight Decay Regularization," in *International Conference on Learning Representations*, 2019.

[20] M. Heusel, H. Ramsauer, T. Unterthiner, B. Nessler, and S. Hochreiter, "GANs Trained by a Two Time-Scale Update Rule Converge to a Local Nash Equilibrium," in *Advances in Neural Information Processing Systems*, 2017.

[21] T. Kynkaanniemi, T. Karras, S. Laine, J. Lehtinen, and T. Aila, "Improved Precision and Recall Metric for Assessing Generative Models," in *Advances in Neural Information Processing Systems*, 2019.

[22] T. Dzanic, K. Shah, and F. Witherden, "Fourier Spectrum Discrepancies in Deep Network Generated

Images," in *Advances in Neural Information Processing Systems*, 2020, pp. 3022–3032.